\documentclass[manuscript,screen,]{acmart}

\renewcommand\footnotetextcopyrightpermission[1]{} 
\usepackage{siunitx} 
\usepackage{graphicx} 
\usepackage{natbib} 
\usepackage{amsmath} 
\usepackage{hyperref}
\usepackage{pdflscape}
\AtBeginDocument{%
  \providecommand\BibTeX{{%
    \normalfont B\kern-0.5em{\scshape i\kern-0.25em b}\kern-0.8em\TeX}}}

\setcopyright{acmcopyright}
\copyrightyear{XXXX}
\acmYear{XXXX}
\acmDOI{10.1145/xxxxxxxxxx}

\begin{document}
\fancyfoot{}
\title{Gaze-based intention estimation: principles, methodologies, and applications in HRI}

\author{Anna Belardinelli}
\email{anna.belardinelli@honda-ri.de}
\orcid{https://orcid.org/0000-0002-0266-3305}
\affiliation{%
  \institution{Honda Research Institute Europe}
  \streetaddress{Carl-Legien-Str.30}
  \city{Offenbach}
  \country{Germany}
  \postcode{63073}
}



\begin{abstract}
Intention prediction has become a relevant field of research in Human-Machine and Human-Robot Interaction. Indeed, any artificial system (co)-operating with and along humans, designed to assist and coordinate its actions with a human partner, {would benefit from first inferring} the human's current intention. To spare the user the cognitive burden of explicitly uttering their goals, this inference relies mostly on behavioral cues deemed indicative of the current action. 
 It has been long known that eye movements are highly anticipatory of the single steps unfolding during a task, hence they can serve as {a very early and reliable behavioural} cue for intention recognition.  { This review aims to draw a line between insights in the psychological literature on visuomotor control and relevant applications of gaze-based intention recognition in technical domains, with a focus on teleoperated and assistive robotic systems. Starting from the cognitive principles underlying the relationship between intentions, eye movements, and action, the use of  eye  tracking  and  gaze-based  models  for  intent  recognition in Human-Robot Interaction is considered, with prevalent methodologies  and  their diverse applications. Finally, special consideration is given to relevant human factors issues and current limitations to be factored in when designing such systems.}
\end{abstract}



\maketitle
\thispagestyle{empty}
\section{Introduction}
\begin{figure}[ht]
    \centering
\includegraphics[width=0.99\textwidth]{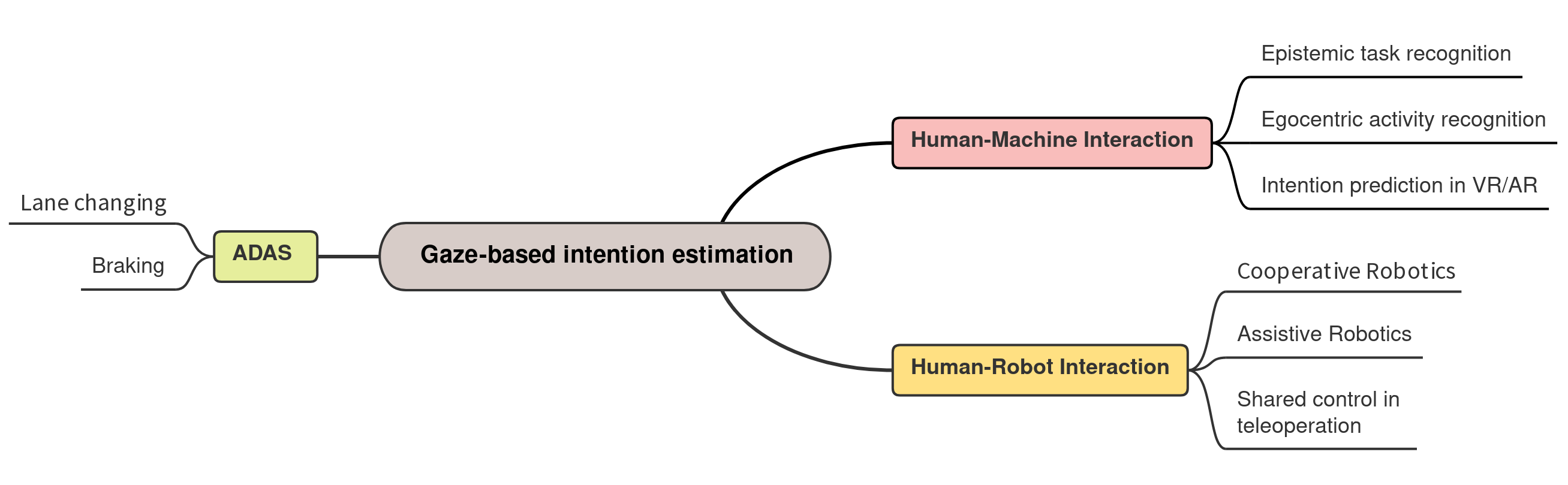}
\caption{Fields and sub-fields of application for gaze-based intention estimation, as surveyed in this review.}
\label{fig:appscheme}
\end{figure}

Intention recognition algorithms are currently at the heart of fast developing technologies involving the interaction of human and artificial systems, may these be teleoperated robotic arms, driving assistance systems, robotic wheelchairs or social robots acting in an environment shared with humans \citep{Vernon2016}. These systems usually rely on a number of machine learning techniques for early {estimation} of the ongoing or upcoming action, typically basing on partially observed time series of behavioral cues. Further, as small and cheap wearable cameras are becoming commonplace, a privileged ego-centric vantage point can be used to effectively observe how intentions evolve into action.
In recent years, it has become clear that {a conveniently}  natural and timely way to predict intention in many applications is to use gaze and eye movements. Considering intentions in physical actions, for example, in light of the need to cope with sensorymotor delays \citep{Miall2002}, gaze control itself in task-based scenarios can be considered as inherently predictive of a number of action-relevant aspects. Indeed, in moving our eyes we make use of knowledge- and sensorymotor-based experience and memory \citep{Henderson2017,Hayhoe2017,Fiehler2019} to quickly retrieve the information needed to plan limb motion. We further use our eyes to initiate and supervise action during its whole execution.  Such gaze cues (among others) are also used by other humans to estimate our intentions and possibly help with those or react accordingly \citep{Emery2000,Tomasello2005,Huang2015}. Gaze is indeed a strong cue in social interaction and it has been often used also in Human-Robot interaction (HRI)  \citep{Admoni2017}.  These considerations show that intention recognition is a highly cognitive task and hence it constitutes a critical feature for intelligent cooperative systems. In general, as pointed out by \citeN{Bulling2014}, situation- or context-aware systems have been developed to proactively help users, but more attention has been paid to the environmental context, somehow neglecting, until recently, the cognitive context of the user.
Taking this into account, for example by inferring intentions, would push cognition-aware computing forward. {Indeed, as argued by \citeN{Zhang2020}, while robotic systems currently relying by design on human-centered AI are increasingly becoming pervasive in human social environments, such agents need to perceive and leverage the human gaze to better understand and interact with humans. As for other cognitive functionalities and models in HRI, one can approach gaze-based intention recognition in a purely data-driven way or in a more hypothesis- or model-driven way \cite{Tsotsos2019}, with the former often preferred lately in applicative fields. Gaze data indeed 
have currently become very easy to collect in large amounts for any specific application one might need, yet here we would like to make the point that a better comprehension of the human visuomotor system could go a long-way also in optimizing such data harnessing. }
In this respect, although machine learning approaches can be used in an agnostic way to learn relevant associations between gaze and intentions, knowing how intentions and actions emerge and coordinate in the human mind through the eyes interplay  might help identify relevant features and design better systems depending on the context and activity of interest. {Since this background knowledge is often missing or not systematized sufficiently in current technical systems}, before surveying the literature on gaze-based intention recognition, we first make an excursus reviewing cognitive science studies on how intentions are formulated, what is their function in action production, and how the gaze and visual systems facilitate this connection in an integrated way, underlining the predictive nature of eye movements in natural eye-hand coordination. {We further consider what gaze features are usually considered, how they are computed, and what kind of insight they can deliver while also looking at which predictive models have been mostly adopted. This also can depend on the  technical fields in which intention recognition has found relevant application (see Figure \ref{fig:appscheme} for a summary). In recent years the use of gaze tracking devices for inferring the user intention has increased but a wide-spread, multi-purpose use is still hampered, that is, intentions have been considered both either as overarching activities (e.g., reading, browsing, see Section \ref{subsec:HMI}) or at the level of primitive actions (e.g., pick, place, or brake, as in Sections \ref{subsec:HRI} and \ref{subsec:ADAS}) but deliberate manipulation intentions of higher semantic level seem more elusive and hard to capture. In this respect, we finally consider a number of limitations, challenges, and open related issues that need to be taken into account in order for intention recognition systems to be more effective and acceptable to human users.}

\section{Intentions, gaze and action}

By surveying the technical literature a number of terms occur that {appear to be} concerned with the same concept as intention recognition: intention estimation, intention prediction, intention inference, action prediction, activity recognition, task prediction,  and, in some cases, plan recognition.
All of them assume that there is a purposeful, deliberate behaviour displayed by a human agent that can be recognized/predicted  by an artificial system on the basis of different observable cues. These cues range from body motion as externally tracked by cameras, to movements detected by body-worn inertial units or mouse/joystick controllers, to signals from myographic activity or brain-computer interfaces, to gaze control, again as observed from external cameras or by egocentric, head mounted eyetracking devices. For the purpose of plan recognition, these cues do not allow to look too far ahead in the future and also a kind of Theory of Mind would be needed, along with a planner \citep{Baker2014}, hence we will not review such systems. As pointed out by \citeN{Kong2018}, even though at times the terms are still not so clearly differentiated, \textit{action recognition} is usually done offline and when the action to recognize has been completed, while \textit{action prediction} relies on preliminary observation of the above mentioned cues to estimate (possibly online) the ongoing action. {Still, this distinction is hardly applicable when speaking of intentions. As reviewed in the following section, intentions are hidden, unobservable states, which first can be inferred when translated into a motor behavior. In this sense, intentions are always recognized or inferred \textit{post-hoc}, since they have been already formed, while the corresponding actions might still be predicted before completion. In this sense, we will consider intention estimation, prediction, recognition, and inference as synonyms here, while using the former as the more generic term. }

For the reasons stated above this review focuses on gaze as the paramount cue for {intention estimation, which can help estimating the intention even before the manual action has started.  In this perspective, what is an intention? What is its relation to action? How is this relation mediated by our visual and gaze control systems?} \\

\subsection{From intentions to actions}
Intentions as mental states have long been considered as a key element in philosophical action theory.
The philosopher Dennett proposed the \textit{intentional stance} as the foundational perspective to predict others' behaviour and actions \citep{Dennett1989}.  Ascribing to any rational agent beliefs, desires, and intentions, we are able to predict what they will do to achieve those intentions.
The concept of intention is indeed strongly related to those of action generation and agency within the field of motor cognition and is at the core of action explanation \citep{Pacherie2012}. Early approaches \citep{Davidson1980,Goldman1970} considered  only beliefs and desires as causal antecedents, more specifically as mental causes of actions. However, action can be considered as a spectrum going from low-level motor primitives to carefully  planned, deliberate actions. In this perspective, intentions were introduced, beside beliefs and desires, to account for future-directed intentions followed by no immediate action, as well as for instinctively, almost automatically performed actions that still denote a minimal purpose  \citep{Bratman1987}. Following this consideration, intentions have been assumed to have a number of functions and have been characterized according to their temporal relation to the consequent action. Specifically, before any actual action is initiated, intentions prompt plans and coordinate the plan activities in time \citep{Bratman1987}, while they further play the three critical functions of initiating, guiding and controlling actions up to their completion \citep{Brand1984,Mele1992}. These distinctions have led to speaking of prior intentions vs. intentions-in-action \citep{Searle1983} or future-directed vs. present-directed intentions \citep{Bratman1987} or distal vs. proximal intentions \citep{Mele1992}. 

\begin{figure}[ht]
    \centering
\includegraphics[width=0.95\textwidth]{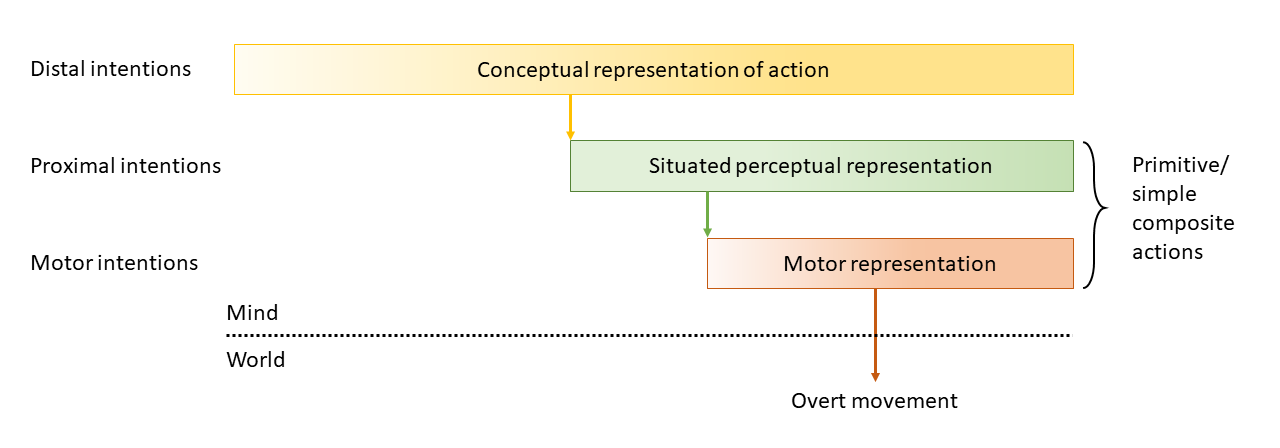}
\caption{A graphical representation of the cascade of intention levels, from abstract distal intentions to motor intentions directly translating into motor actions in a situated way (redrawn from \cite{Pacherie2008}).}
\label{fig:Intentions}
\end{figure}

In this line of thinking,  \citeN{Pacherie2008} has proposed an even finer characterization: D-intentions, P-intentions, and M-intentions. This distinction relies on the consideration that distal intentions do not terminate their function once proximal intentions take their monitoring and guidance role. The three proposed intentions (distal, proximal and motor) do instead work in parallel on different levels of representation content and at different time scales. In this framework, D-intentions define which goal to pursue and can be flexible regarding the time and context, leaving the planned action parameters and constrains rather unspecified. P-intentions receive an action plan from D-intentions and trigger off the start of the action, situating it in a specific time and context, relying on explicit perception, situation awareness and thought. P-intentions, still, may act on higher level action descriptions while M-intentions work with specific action primitives and motor representations (see Figure \ref{fig:Intentions}). The three levels of course operate on different time scales, being in the range of minutes and up for D-intentions, of few seconds for P-intentions and of hundreds of milliseconds for M-intentions.\\

These formalizations show that action emerges from plans and motor programs combined in a functional architecture relying on forward and inverse models. This means that any behavioral observation that can be related to the formulation of a motor plan entails the potential to reveal the action associated with the current proximal intention. In this sense, although a number of approaches have relied on motion cues to infer intentions  \citep{Yu2005,Javdani2015,Hauser2013,Aarno08,Tanwani2017,Tahboub2006}, an even earlier indicator, one which already displays during action preparation, is gaze control. Why is this the case? How do the visual and gaze systems mediate the transformation of an intention into motor action?

\subsection{Gaze-based prediction and eye-hand coordination in natural tasks}\label{subsec:eye-hand}

Task strongly influences attention and consequently shapes eye movement patterns: in an early eye tracking experiment, \citeN{Yarbus1967} showed that presenting the same picture to subjects, asking them every time a different question (e.g., estimate the age of the pictured people, estimate their wealth), produced reliably different scanpaths.

These days most of our visual activity might be concerned with passively processing  information from texts, images, videos, and other media. Yet, our vision organs (and cognition altogether) have actually developed and evolved in a moving organism that needed to act in the environment in order to adapt and survive. Even during a simple visual search, our  performance is affected by action intentions and possibilities \citep{Bekkering02, Bub13}. At a theoretical level this was already postulated by the premotor theory of attention \citep{Rizzolatti87}, arguing that spatial attention emerges as enhanced spatial processing at locations of possible actions, suggesting a tight relation and shared mechanisms for attention and action, and specifically for selection-for-perception and selection-for-action \citep{Deubel98}.  When we intend to act (or we need to react), that is the very moment our information-seeking perceptual behaviour must be most effectively tuned on the final goal. Action is indeed mostly sequential (we hardly coordinate movements with different goals at the same time) and it is relatively slowed down by inertia in execution. Thus, in order to maximize the probability of action success, motor planning needs to act on timely and certain information provided by perception. This corresponds to a visuomotor behaviour policy that strives to maximize the information gain and minimize uncertainty or surprise \citep{Friston15,Itti09}.
To do that, the visual system needs to rely on previous experience and knowledge, so to anticipate the critical information  to supply while preparing for and controlling a particular goal-directed action, in accordance with the current motor and subtask needs. Although this is apparent also in locomotion and navigation tasks \citep{Einhaeuser07,Werneke12}, it can be appreciated best in the interaction with objects. 
The advent of head-mounted eye-trackers in the last two decades has made possible the study of eye-hand coordination in an ecological way, with subjects performing daily tasks like making a sandwich \citep{Hayhoe2003}, preparing tea \citep{Land1999}, walking around \citep{Hart2012} or practicing sports \citep[for a thorough review see ][]{Land2009}. Specifically considering eye-hand coordination, \cite{Johansson2001} looked at the timing correlations in the behaviour of gaze and hand when a simple wooden block had to be grasped, lifted and carried to make contact to a target position, meanwhile avoiding a protruding obstacle. 

A number of lessons have been learned from these studies, especially considering structured manual interaction with objects:   
\begin{itemize}

\item for the most part only task-relevant objects are targeted by the gaze \citep{Land2006}; 
\item objects are glanced at mostly right before they are needed for the task -'look-ahead' and 'just-in-time' fixations \citep{Pelz2001,Hayhoe2003}-  and are further fixated until right before the hand grasps them or the task is completed. That is, the eyes might move ahead to the next sub-goal even if the motor task is not completely accomplished;
\item the hand or the objects carried in the hand are rarely look at \citep{Johansson2001}. In general, an anticipation of the eye with respect to the hand (eye-hand span) of 0.5 up to 1 s is observed, which {demonstrating that gaze can be used to predict the next manipulation step.} 
\end{itemize}

This suggests that the way oculomotor and hand motor plans unfold during the execution of an eye-hand coordination (manipulation) task and with respect to the involved objects offers a substantial insight into the cognitive structure of action representation and planning, as orchestrated by intentions. Once we are set to achieve a specific goal, directions for a parallel and tightly interleaved execution of two behaviors (information seeking and object manipulation) are formed and delivered to the gaze and visual system, on the one hand, and to the motor system on the other. 

\begin{figure}[h]
    \centering
\includegraphics[width=0.95\textwidth]{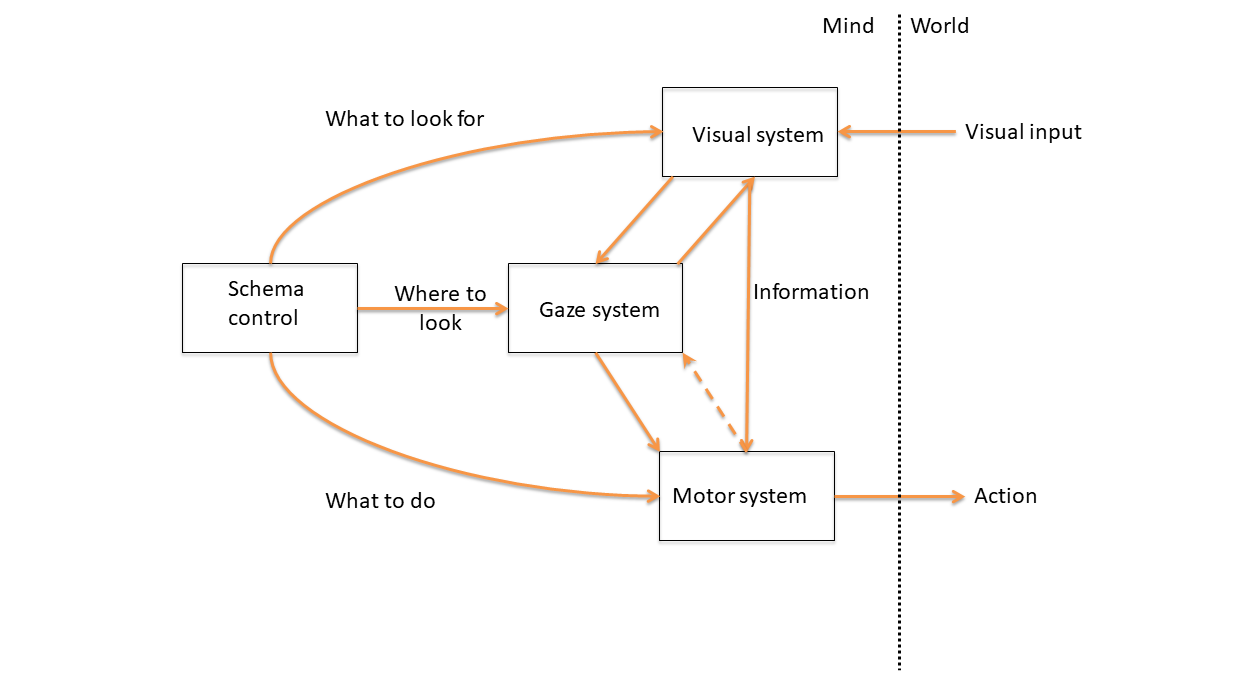}
\caption{Information exchange between the functional modules regulating perception and action: the relation between gaze control, visual processing and action control is orchestrated by the schema system which sets the agenda for locations to look at, objects to look for and movements to plan. The solid arrows show the information flow as sketched by \cite{Land09}, here redrawn; the dashed arrow represents a further information exchange suggested here and demonstrated in \citep{Brouwer2009,Belardinelli2016a,Belardinelli2016b}.}
\label{fig:landscheme}
\end{figure}

\citeN{Land09} proposes that four systems are involved in this process (see Fig. \ref{fig:landscheme}): 
\begin{enumerate}
\item The \textit{schema control} 
 sets the goal and the coarse plan to achieve it by defining the agenda of the necessary steps in the task sequence. Specifically, general directions about the objects of each subtask, where to find them, and the motor primitives necessary for the interaction  need to be communicated to the other systems down the line. 
\item  The \textit{gaze system} encompasses control of each body part movement aimed at directing our eyes to the most likely locations where the objects to be acted upon can be found. Hence, body/trunk movements, head movements, and eye movements are all orchestrated by the gaze system\footnote{Even though one might argue that these movements are actually controlled by the motor system, in this case these exploratory movements have an epistemic value (reducing the uncertainty) while the movements actually performing the action can be considered having a pragmatic value in reaching the goal state \citep{Friston15}.}. 
\item The \textit{visual system} is responsible for the processing of the visual input and the recognition of the objects and context of the visual scene. It also monitors that the action unfolds successfully, achieving the goal of each task. 
\item The \textit{motor system} {physically executes} the intended action, combining the available motor primitives (like reaching and grasping) with the contingent object configuration, as informed from the visual system. 

\end{enumerate}

{Clearly, these systems must work in tight concert with one another and } information must be exchanged, especially between the gaze and visual system and between these two and the motor system. If, for instance, the object we are looking for is not detected by the visual system in the current field of view, the gaze system must know that further exploration is needed. Otherwise, if the target of the action was retrieved, the motor system must know where in the field of view  and with respect to the rest of the body this is and how the object might be grasped. These information exchanges are represented in Fig. \ref{fig:landscheme} by the solid arrows between the different systems.

The activity of the schema control system is behaviorally not observable, but its relation to the visual, gaze, and motor systems allows to draw inferences about its agenda by observing  fixation locations, scanpaths, and possibly hand movements and their temporal synchronization during a task.
Not just what object is going to be grasped or used next can be inferred by fixations ahead of the hand reaching it. A number of studies have shown how the task, the intended grasp and the user's experience with the object have an effect on fixations even before the hand reaches for the object \citep{Brouwer2009,Belardinelli2015,Belardinelli2016a,Belardinelli2016b}.

When planning a reaching to grasp movement, the eyes usually target the index finger contact location since this is usually the first finger to make contact with the object \citep{Brouwer2009,Cavina2013}. 
In general,  the relevant affording points on the object are visually targeted, if possible also in parallel, hence either covertly or targeting a fixation point in the middle. In this way, a kind of attentional landscape is formed, enhancing perceptual processing in relation to the spatio-temporal action requirements \citep{Baldauf2010}.
 That is, by observing the visual exploration of objects and the ensuing manipulation  under different tasks and conditions, the diverse information requirements that are needed to plan and execute each task become apparent.

What these works additionally suggest is that a further arrow is missing from the schema above, namely information flowing from the motor system to the gaze system. That is, only once the motor plan for the appropriate grasp has been selected,  the eyes can move to the most convenient location to guide the grasp. This exchange is denoted by the dashed arrow in Fig. \ref{fig:landscheme}.

The fact that the eyes precede the hand and satisfy its informational needs while the motor system communicates its kinematic needs to the gaze, all prior to or while initiating action execution, demonstrates how perception and action are substantially embodied and integrated processes, relying on learning and anticipation of the intended sensorimotor goal states.\\

{What seen so far concerns the cognitive basis of visual attention and oculomotor control during grasping and manipulation.}
While still engaging eye-hand (and eye-foot) coordination, separate considerations must be done for a more complex activity such as driving, {which further is the domain of a number of applications that need to rely on timely intention estimation for human-machine cooperation}.
Gaze control, as seen above, works according to spatial predictions about where to find relevant information in the current scene \citep{Henderson2017}, whereas these predictions are often made on the basis of episodic experience or knowledge. Coupled with top-down endogenous attention, this predictive ability makes vision particularly efficient and capable of focusing on relevant information within complex scenes in a matter of hundreds of milliseconds.
Yet, at the same time we must be able to deal with unexpected stimuli, especially those not relevant to our current task but whose detection is critical to our and others’ safety. This is particularly an issue, of course, when acting within dynamic environments, where an erroneous prediction requires a prompt reaction.
Driving takes place in such environments and the scene to be scanned is usually very rich with information with different degrees of relevance. 
Driving is also inherently multitasking: it is not indeed just about steering the vehicle in the intended direction but it is made of different parallel subtasks (collision avoidance, navigation planning, lane keeping,...), each running with time-changing priorities, hence requiring continuous (re)scheduling of attentional resources allocation \citep{Hayhoe2014,Rosner2019}. That is, if in tasks with a relative sequential structure, the gaze behaviour can be considered as a running commentary of the current cognitive agenda and can reveal to a certain degree the current subtask, fixation choices in a multi-tasking behaviour need to be made not according to a linear unfolding of the task, but depending on competing modules, which are to be periodically updated with fresh information, especially when the world state cannot be assumed to be static.
To complicate the picture further, scanning behaviours might change depending on the context (urban vs. highway or country driving), on the current manoeuvre to be undertaken (overtaking or crossing an intersection), and on the current speed. 
These factors not only influence gaze location, but the whole visuo-motor control system. {Following a bend, the eyes fixate on the tangent point of the line of sight to the lane edge \citep{Land1994,Mars2008}, leading the steering hand by about 0.8 s.} 
In urban traffic, the gaze might shift every .5 s, alternating between the car in the front, lane edges, pedestrians, road signs, traffic lights or generic obstacles \citep{Land2009}. Still, it is hard to make generalizations on gaze distribution independently of the situation at hand. Moreover, it is known that the higher the cognitive load the more attention concentrates on foveal vision, to the expenses of peripheral vision \citep{Crundall2005}.\\
Particular attention has been given to gaze behavior during lane changing, a situation that requires tight coordination of situation awareness, decision making and motor control because of the interplay with other moving vehicles. \citeN{Salvucci2002} looked at gaze dwell time  {in such case 
and found that if in the beginning (lane keeping) the gaze mostly lingers on the start lane, when changing from right to left lane, as the change time approaches more time is spent on the mirror to check incoming vehicles and on the end lane. 
When changing from left to right instead, glances at the mirror increase shortly before lane change, but during and after lane change the gaze spends most of the time on the destination lane. The authors remark how the gaze, anticipating the manoeuvre, within 1-2 s of the onset of lane change is already concentrated on the destination lane, although the vehicle is actually still entirely in the start lane.
All these results show that a number of factors can influence the way our gaze is controlled when driving \citep[see ][ for a thorough review]{Kotseruba2021}. }
This could make the assessment of the driver's intentions  particularly tricky, since the process that determined a series of fixations (intentional retrieval of information or reflexive capture of attention) might not be easily pinpointed. Still a combination of behavioural cues could provide a much richer picture.\\

As a final remark, even if intentions, as we have seen, can have multiple temporal granularities, ranging from distal to proximal to primitive motor intentions, the studies above however show that 'just-in-time' fixations in physical actions denote an action planning policy according to which visual information about the immediate next action is retrieved right ahead (i.e., in the order of hundreds of milliseconds before starting the action). After that, the eyes then stay anchored on the object of interest as long as visual guidance for manipulation is needed. In driving scenarios, although the time span between the eyes collecting information for a manoeuvre and the hand initiating the manoeuvre is usually longer, it still stays in the range of 1 to 5 seconds.  Such time ranges hardly help retrieving long-term, distal intentions. 

{For these reasons in the following we concentrate preeminently on models and systems handling proximal and motor intention recognition, driving single, often primitive, actions. We aim at showing how indeed such cases represent thus far the most compelling use of gaze for intention recognition and how such intentions can be easily retrieved by means of easily computed features, as presented in the next section.}

\section{Gaze features and models for intention prediction}

 Before looking at how the gaze has been incorporated in different application domains, we give a brief overview of the gaze and eye movements features and measures which are mostly used and of the machine learning models that can make the most out of them, depending on the goal. Although there are a number of studies working with electro-oculography  \citep[see ][ for a review]{Bulling2010} or using just optical flow features caused by head movement  \citep[e.g., ][]{Kitani2011} or gaze direction extracted by external cameras \citep[e.g., ][]{Dermy2019}, we will focus here mostly on gaze- and eye tracking data collected with head-mounted devices, as this technique provides a more accurate position of the gaze and localizes it with respect to the scene, hence linking the eye movement patterns to the specific object or location of interest at any time. 
 
 Different abbreviations which will be used in the following are reported in Table \ref{tab:abbrev}.
\begin{table}[]
\begin{tabular}{|l|l|l|l|ll}
\hline
{ \textbf{Acronym}} &
  { \textbf{Meaning}} &
  { \textbf{Acronym}} &
  { \textbf{Meaning}} &
  \multicolumn{1}{l|}{{ \textbf{Acronym}}} &
  \multicolumn{1}{l|}{{ \textbf{Meaning}}} \\ \hline
\textbf{ADAS} &
  \begin{tabular}[c]{@{}l@{}}Advanced Driving \\ Assisting Systems\end{tabular} &
  \textbf{GMM} &
  \begin{tabular}[c]{@{}l@{}}Gaussian Mixture\\ Model\end{tabular} &
  \multicolumn{1}{l|}{\textbf{POMDP}} &
  \multicolumn{1}{l|}{\begin{tabular}[c]{@{}l@{}}Partially Observable \\ Markov  Decision \\ Process\end{tabular}} \\ \hline
\textbf{AOI} &
  Areas Of Interest &
  \textbf{HMI} &
  \begin{tabular}[c]{@{}l@{}}Human-Machine \\ Interaction\end{tabular} &
  \multicolumn{1}{l|}{\textbf{POR}} &
  \multicolumn{1}{l|}{Point Of Regard} \\ \hline
\textbf{AUC} &
  \begin{tabular}[c]{@{}l@{}}Area-Under-\\ The-Curve\end{tabular} &
  \textbf{HRI} &
  \begin{tabular}[c]{@{}l@{}}Human-Robot\\ Interaction\end{tabular} &
  \multicolumn{1}{l|}{\textbf{RBF}} &
  \multicolumn{1}{l|}{Radial Basis Function} \\ \hline
\textbf{CRF} &
  \begin{tabular}[c]{@{}l@{}}Conditional Random\\ Field\end{tabular} &
  \textbf{k-NN} &
  \begin{tabular}[c]{@{}l@{}}k-Nearest \\ Neighbors\end{tabular} &
  \multicolumn{1}{l|}{\textbf{(R)NN}} &
  \multicolumn{1}{l|}{\begin{tabular}[c]{@{}l@{}}(Recurrent) \\ Neural Networks\end{tabular}} \\ \hline
\textbf{(D)BN} &
  \begin{tabular}[c]{@{}l@{}}(Dynamic) Bayes \\ Network\end{tabular} &
  \textbf{LDA} &
  \begin{tabular}[c]{@{}l@{}}Linear Discriminant \\ Analysis\end{tabular} &
  \multicolumn{1}{l|}{\textbf{SVM}} &
  \multicolumn{1}{l|}{\begin{tabular}[c]{@{}l@{}}Support Vector \\ Machine\end{tabular}} \\ \hline
\textbf{DTW} &
  \begin{tabular}[c]{@{}l@{}}Dynamic Time \\ Warping\end{tabular} &
  \textbf{LSTM} &
  \begin{tabular}[c]{@{}l@{}}Long Short-Term \\ Memory\end{tabular} &
  \multicolumn{1}{l|}{\textbf{VR}} &
  \multicolumn{1}{l|}{Virtual Reality} \\ \hline
\textbf{(G)HMM} &
  \begin{tabular}[c]{@{}l@{}}(Gaussian) Hidden \\ Markov Models\end{tabular} &
  \textbf{MAP} &
  \begin{tabular}[c]{@{}l@{}}Maximum A \\ Posteriori\end{tabular} &
   &
   \\ \cline{1-4}
\end{tabular}
\caption{Abbreviations used in the text.}
\label{tab:abbrev}
\end{table}

 \subsection{Gaze features: position, movement, sequence and semantic features}
 Without referring to any particular hardware, every eye tracking device delivers as raw data the eye(s) position as 2D coordinates on a calibrated surface or in the scene in terms of frame coordinates with respect to the streaming of a head-mounted scene camera. For many applications considering physical interaction with objects in the world, it might be necessary to localize the point of regard (POR) in 3D and specific methods have been proposed for that, often relying on fiducial markers \citep[e.g., ][]{Huang2016,Fathaliyan2018,Takahashi2018} but new solutions are trying to do without \citep{Weber2018,Liu2020}. 
 
 In any case, once gaze coordinates are obtained, these can be used as such or to compute gaze events of interest, usually fixations and saccades, by means of specific detectors. Fixations are periods of time in which the gaze is relatively stationary on a certain point, while saccades denote fast displacements of the POR between fixations. Fixations are generally in the order of 200-300 ms while saccades can take between 30 and 80 ms. The reader is referred to \citeN{Holmqvist2011} for a more thorough classification of eye movements.
 During saccades the visual input is blurred and hardly any visual processing can take place, hence often just fixations are considered when analyzing eye data, considering visual intake to happen only in those moments. To this end, fixations can be detected via dispersion- or velocity-based algorithms \citep{Salvucci2000}, which are yet based on predefined thresholds.  {In more recent years,  some machine learning-based event detection schemes have been proposed, which achieve a higher accuracy \citep{Vanderlans2011,Tafaj2012,Zemblys2018}}.   Still, when a system is working online, as needed for real-time prediction, it can make sense to use all gaze samples in order not to have an online fixation detector processing data before intention classification \citep{Coutrot2018,Fuchs2021}. Simply using the PORs indeed accounts for both the gaze location and temporal distribution on the object of interest \citep[as in ][]{Keshava2020}.
 
 For activity recognition or task prediction, in some cases the pattern of gaze shifts can be used independently of the underlying visual input  \citep[as in, e.g., ][]{Ogaki2012,Bulling2010}. In this sense, \cite{Srivastava2018} make a distinction on these kind of features: as low-level features are considered duration and position statistics of just single fixations/saccades, while statistics of shape-based and distance-based patterns (quantized as strings) are considered as mid-level features. Such patterns are indeed differently distributed in desktop activities, for example \citep{Ogaki2012,Srivastava2018}.
 Often,  fixation and saccade statistics (such as fixation duration or saccade amplitude means and variances) along with the statistics of image features at fixation landing points (e.g., Histogram-Of-Gradients, saliency or gist features) are used to infer the current epistemic task \citep[see ][for a review]{Boisvert2016}.
 
However, when considering the prediction of a physical interaction,  higher level - somehow more semantic - features need to be considered. Although gaze distribution on images or recurrent pattern of directional saccadic shifts can be revealing of specific repetitive activities, these features are hardly relatable to a single action intention (although, they can be revealing at least of the intent to interact  \cite{David-John2021}). The next proximal intention is indeed often revealed by the current object of interest and even by a specific affordance on that object \citep{Belardinelli2015,Belardinelli2016b}, independently of their position in the camera image or of specific low-level features at fixation points. The relevant objects in the scene should hence be defined as Areas-Of-Interest (AOI) and  any fixation within these areas gets considered for processing, with related statistics. In this case, AOI-specific measures can be derived, e.g. the number of fixations within each AOI, the time of entry in each AOI and the overall dwell time. More importantly, considering the temporal evolution of intentions and actions, sequences of glanced AOIs can be used as time series of discrete symbols, revealing the shifting of attention between different objects relevant for the current activity, as for example in \citeN{Fathaliyan2018}. To account for oculomotor inaccuracy and for the strategic use of parafoveal vision, in some cases, instead of assigning at any time the gaze in full to just one object, the POR is represented as a Gaussian distribution \citep[e.g, as in ][]{Admoni2016,Fuchs2021} possibly intersecting different AOIs or time bins of gaze counts are considered in which a discrete distribution over AOIs is computed \citep{Lengyal2021}.\\
 AOIs are often used as semantic inertial areas within vehicles. While estimating the current object of interest in a dynamic scene with ego-motion is technically quite complex \citep[but see ][ for a recent solution]{Jiang2018}, intentions like lane keeping or lane changing can be assessed considering which part of the field of view within the car the driver is currently attending: often the windscreen, the rear and lateral mirrors and the the dashboard are taken as AOIs.
 
 In any context, sequences of AOIs can be considered as strings where typical, recurrent transitions lend themselves nicely to be learnt by Markov models, as discussed in the next subsection.
 
 \subsection{Prediction models}
 
Typically, intention recognition systems feed gaze features considered indicative of the current intent into some classifier outputting the most likely intention. For this purpose often Support Vector Machines (SVM) \citep{Huang2016,Kanan2014}, random forests \citep{Boisvert2016}, logistic regression \cite{David-John2021}, and neural networks \citep{Lethaus2013} have been used.  In some cases, like task prediction in viewing images \citep{Henderson2013} or driving manoeuvres \citep{Lethaus2013}, a Naive Bayes classifier produced a satisfying intention recognition, yet most established approaches use more sophisticated generative models. 
 Indeed, as mentioned earlier, for intention prediction we want to estimate a cognitive state which is \textit{per se} not observable, that is, a hidden state. For this reason intention recognition has often been modeled via probabilistic graphical models, like Dynamic Bayesian Networks and Hidden Markov Models \citep{Tahboub2006,Bader2009,Yi2009,Boccignone2019,Haji2014,Fuchs2021}. In general, if the intention $I$ represents the hidden state and what can be observed (the eye features of interest) is denoted with $E$, intention recognition can be typically defined in Bayesian terms:
\begin{equation}
 P(I|E) = \frac{P(E|I)P(I)}{P(E)}    
\end{equation}

If we have a model $M_I$ for each intention and a sequence of features $E_{0..t}$, inference of the current intention $I_t$ is given by the model which maximizes the posterior of the observed data:

\begin{equation}
 I_t = \underset{I}{\mathrm{argmax}}\ P(M_I|E_{0..t}) = \frac{P(E_{0..t}|M_I)P(M_I)}{P(E_{0..t})}
 \label{eq:prediction}
\end{equation}

A number of recent approaches have also used recurrent neural networks such as Long Short Term Memory (LSTM) \citep{Min2017,Schydlo2018,Gonzalez2019,Wang2020} and Partially Observable Decision Markov Models (POMD) \citep{Admoni2016}, which are again algorithms that deal well with time series of different length.

In general, it makes sense to use models that account for the temporal evolution of the scanpath on the scene. To consider just the current fixation or POR, in fact, would produce a very erratic prediction whenever the gaze lands on anything which is not strictly related to the current intention. Even the approaches relying on discriminative classifiers mentioned above do not consider gaze features at a specific point in time, but often compute some summary statistics within a certain time window. By doing so, a larger time history is taken into consideration, still the temporal information is lost. Generative models also rely on data observed in a certain time window but they do incorporate an internal representation of the task evolution. This makes them also more robust against spurious fixations.
Still, the time window to consider must be carefully chosen in order to capture the time scale of the intentions to be estimated.  For reaching, grasping, and simple manipulations a time window between 0.5 and 2 s is usually delivering best results \citep{Huang2016,Fuchs2021}, while in driving scenarios a longer time frame is usually considered, between 5 and 10 s \citep{Lethaus2013}.\\

Further, while Hidden Markov Models can handle sequences of different length in a relatively flexible way, sometimes time-series need to be reduced to comparable lengths in order for a distance or similarity measure to be computed. This problem has been mostly tackled by dynamic time warping \citep{Fathaliyan2018,Wu2019} or by encoding sequences in a fixed length representation by means of an encoder-decoder LSTM architecture \citep{Schydlo2018}.\\

Finally, a further aspect to consider is that in many situations instead of a classification scheme returning just the winning intention, it is better to have a distribution over all possible intentions. The selection can then happen by means of an argmax function (as shown in Equation \ref{eq:prediction}), still a measure of prediction confidence can be given, for example by taking the difference between the two most likely intentions \citep{Jain2018}. In this way the following assistance can be tuned more or less strongly, depending on how sure the system is about the estimated intention.
{In the following section, we illustrate which of the presented techniques and features have been prevalently used in multiple applicative fields, depending on the use case and nature of the intention to recognize.}

\section{Gaze-based intention {estimation}  in technical domains} \label{sec:gaze_intent}

{As anticipated in Figure \ref{fig:appscheme} there are three main fields where gaze features have been used for intention recognition in a wide sense. In Computer Vision, often intent has been recognized  \textit{a posteriori}, offline, that is, inferring the epistemic task offline from collected data. Although  some more interactive applications could be envisioned for video games, VR or any further application of Extended Reality (XR), mostly in this domain we see intention recognition as activity/task classification, relying on features related to overall fixation distributions. Conversely, when prompt reaction to or support for human real-world activities from robotic systems, such as teleoperated arms or driver-assistance systems, AOIs related to the object of interest and time-series acquire prominence, with the earliness of recognition being an important parameter. We survey here some paradigmatic examples of systems across the three domains.  }

\subsection{Computer Vision and Human-Computer Interfaces}\label{subsec:HMI}
In the last two decades, accompanied by the evolution of eye tracking technology and suitable machine learning algorithms, eye movements have been often used to reverse the Yarbus experiment and infer the task (the asked question)  above chance level  \citep[e.g.,][]{Borji2014,Haji2014,Kanan2014}. Besides Yarbus' questions, most of these approaches have worked with the classification of passive information-seeking tasks performed on static pictures, like counting the instances of a certain item, visual search, scene memorization, reading. Úsually these studies make use of features such as statistics parameters of the fixation duration and saccade amplitude distributions,  number of fixations, transition matrices, fixation proportion and duration on some area of interests (faces, bodies, objects). Popular and effective techniques to compute the probability of a given task given eye movements and possibly their sequence entail Naive Bayes classifiers, HMM, SVM, multivariate pattern analysis and random forests \citep[see for a more complete review][]{Boisvert2016}.

Still, what is recognized (and rarely predicted) in these cases is indeed the general epistemic task rather than a specific intention. Gaze-based intention recognition finds more relevant application when considering tasks involving the physical and manual interaction of the observer with the (dynamic) scene. The largely increased diffusion of wearable cameras has triggered research on daily activities recognition as observed from an egocentric perspective \citep{Yi2009,Fathi2012,Ogaki2012}, hence relying on eye, hand, head and possibly body coordination \citep[see][for a full review]{Nguyen2016}.

Yet these approaches are more concerned with  {overarching} activity recognition rather than with simple action or {proximal/motor} intention recognition{, something particularly useful in VR or in interactive games}. 

\begin{table}[]
{
\begin{tabular}{|p{2.5cm}|l|l|l|l|}
\hline
\textbf{Study} &
  \textbf{Task} &
  \textbf{Features} &
  \textbf{\begin{tabular}[c]{@{}l@{}}Recognition \\ method\end{tabular}} &
  \textbf{\begin{tabular}[c]{@{}l@{}}What is \\recognized\end{tabular}}\\ \hline
\cite{Borji2014} &
  \begin{tabular}[c]{@{}l@{}}Epistemic tasks +\\ Yarbus tasks\end{tabular} &
  \begin{tabular}[c]{@{}l@{}}Fixation density \\ NSS\end{tabular} &
  \begin{tabular}[c]{@{}l@{}}kNN, \\ Boosting \\ classifier\end{tabular}&
  \begin{tabular}[c]{@{}l@{}} Task label \\ (e.g., age guessing)\end{tabular} \\ \hline
\cite{Haji2014} &
  Epistemic tasks &
  Sequences of fixation locations &
  HMM & Task label\\ \hline
\cite{Kanan2014} &
  Epistemic tasks &
  \begin{tabular}[c]{@{}l@{}}Fixation number, \\ duration and Fisher \\ kernel features on location\end{tabular} &
  SVM & Task label \\ \hline
\cite{Boisvert2016} &
  \begin{tabular}[c]{@{}l@{}}Free viewing and \\ epistemic tasks\end{tabular} &
  \begin{tabular}[c]{@{}l@{}}Fixation location, density,\\ HOG, Gist features\end{tabular} &
  \begin{tabular}[c]{@{}l@{}}Random \\ forests\end{tabular} & Task label \\ \hline
\cite{Yi2009} &
  \begin{tabular}[c]{@{}l@{}}Sandwich making\\ (with subtasks)\end{tabular} &
  Fixations, AOI sequences &
  DBN & \begin{tabular}[c]{@{}l@{}}Subtask\\ (off- and online)\end{tabular}\\ \hline
\cite{Fathi2012} &
  Meal preparation &
  \begin{tabular}[c]{@{}l@{}}Gaze location, \\ object-based features\end{tabular} &
  \begin{tabular}[c]{@{}l@{}}SVM + \\ generative\\ model\end{tabular} & \begin{tabular}[c]{@{}l@{}}Subtask\\ (verb + object)\end{tabular} \\ \hline
\cite{Ogaki2012} &
  Desktop actions &
  \begin{tabular}[c]{@{}l@{}}n-gram statistics of eye \\ and ego-motion\end{tabular} &
  SVM & \begin{tabular}[c]{@{}l@{}}Activity label\\ (e.g., browsing)\end{tabular}\\ \hline
\cite{Keshava2020} &
  \begin{tabular}[c]{@{}l@{}}Aligning cubes\\ in VR\end{tabular} &
  POR distribution in AOIs &
  \begin{tabular}[c]{@{}l@{}}Radial-basis\\ SVM \end{tabular} & Type of alignment\\ \hline
\cite{Bader2009} &
  \begin{tabular}[c]{@{}l@{}}Virtual pick and\\ place on screen\end{tabular} &
  \begin{tabular}[c]{@{}l@{}}Fixation location and \\ reactive/proactive behaviour\end{tabular} &
 Bayes Net & picking/placing target \\ \hline
 \cite{David-John2021} &
  \begin{tabular}[c]{@{}l@{}} Ingredient fetching\\ in VR\end{tabular} &
  \begin{tabular}[c]{@{}l@{}}Multiple features \\from gaze dynamics\end{tabular} &
\begin{tabular}[c]{@{}l@{}}Logistic \\regression\end{tabular} & Onset of interaction \\ \hline
\cite{Singh2020} &
  \begin{tabular}[c]{@{}l@{}}Board game \\ \textit{(Ticket to ride)}\end{tabular} &
  \begin{tabular}[c]{@{}l@{}}Number and duration of \\ fixations on AOIs\end{tabular} &
  \begin{tabular}[c]{@{}l@{}}Normalized\\ fixations\\ weights\end{tabular} & \begin{tabular}[c]{@{}l@{}}Route to\\target destination\end{tabular} \\ \hline
\cite{Min2017} &
  \begin{tabular}[c]{@{}l@{}}Open-world\\ educational\\ videogame\end{tabular} &
  Fixations on virtual objects &
  \begin{tabular}[c]{@{}l@{}}LSTM +\\ CRF\end{tabular} & \begin{tabular}[c]{@{}l@{}}Goal/Task\\ in the gameworld\end{tabular} \\ \hline
\end{tabular}
\caption{Approaches to task and intention recognition in human-machine interaction.}
\label{tab:HMI}
}
\end{table}

Considering for example simple manual actions in Virtual Reality, \citeN{Keshava2020} have recently shown that the sheer proportion of PORs on different AOIs constitutes a reliable signature, discriminating among different tasks. Gaze-based intention recognition in manipulation of virtual objects on a large screen on a tabletop had already been proposed in \citeN{Bader2009} 
in a pick and place task. 
A Bayes net was used to predict intentions on the basis of proactive fixations in order to improve the robustness of the input device or increase interaction possibilities. 
On the other end of the spectrum, an approach aimed at longer term plan recognition was presented by \citeN{Singh2020}. In a digitized version of a board game considering routes in a graph form, the number and duration of fixations on any AOIs determined scores used as priors over possible trajectories leading to distal intentions. 
The authors show that by means of such gaze priors both proximal and distal intentions could be inferred earlier and more accurately.\\
In the field of digital games plan and goal recognition by means of gaze would help tailor the gameplay on the intentions of the player, showing personalized narratives or interaction with the non-player characters. \cite{Min2017} showed how combining gaze and game events in an open-world game helped recognize the player's goal with an LSTM architecture better than only with game event features.

The studies presented in this section are summed up in Table \ref{tab:HMI}.
All these studies are of interest for the approaches they present with respect to intent recognition, still most set ups do not involve cooperative, shared tasks, nor the user trying to control a physical system {with effects in the real world, where hence timely intention recognition is critical for smooth HRI}. Works related to this theme are reviewed next.

\subsection{Human-Robot Interaction}\label{subsec:HRI}
Intention recognition finds probably its most appropriate domain of application in robotics  since in this field a physical system can become aware of what goal the human is pursuing and proactively provide its support, by acting in the same environment (Section \ref{subsec:soc_rob}) or in a remote/not accessible one (Section \ref{subsec:SA}).

\subsubsection{Human-Robot Cooperation and Social Robotics}\label{subsec:soc_rob}

For human-robot collaboration often the robot partner is aware of the activity context and for effective cooperation it just needs to detect the current action intention of the human partner to help them with it.
In one of the first attempts at using eye tracking in a human-robot cooperative scenario, \citeN{Sakita2004} proposed a system helping a human in an assembly task. The robot would hand the next piece to assemble if it had confidently detected that as being the next step intended by the user or directly execute the step in parallel with the human. The gaze is apparently localized in the real world setup but very few details were given as to the implementation and accuracy.

In most cases, indeed, the user's gaze is mapped on the 2D scene acquired by the eye tracker frontal camera. This is rather inconvenient when investigating interaction with real 3D objects. To tackle this issue, \citeN{Fathaliyan2018} propose a method to localize gaze on 3D objects by projecting the gaze vector on point cloud representations of objects manipulated by a person preparing a powdered drink. 3D heat maps were produced, displaying the most gazed locations on each object depending on the performed subtask. 

In some cases, the gaze is used to identify the object the user intends to select, given a predefined action. For example, 
\citeN{Huang2016} proposed a method for anticipatory control under which  the robot predicts the intent of the human user and plans ahead of an explicit command. In the considered task, a robotic arm prepares a smoothie by picking the ingredients vocally selected  by a human user looking at an illustrated list. 
By means of eye tracking the robot anticipates picking the intended ingredient, hence speeding up task completion. The intent recognition relies on the same features and training set as in a previous work \citep{Huang2015}, where an SVM was fed a feature vector for each ingredient, including the number of glances, duration of the first glance, total duration, and whether it was the most recently glanced item as predictors of the currently intended ingredient. Although such an approach is simple and effective, here the human customer is using their gaze purposely and solely for target selection, with no parallel visuomotor control task that could yield spurious fixations. 
\begin{table}[]
{
\begin{tabular}{|p{2.5cm}|l|l|l|l|}
\hline
\textbf{Study} &
  \textbf{Task} &
  \textbf{Eye Features} &
  \textbf{\begin{tabular}[c]{@{}l@{}}Recognition \\ method\end{tabular}} &
  \textbf{\begin{tabular}[c]{@{}l@{}}What is \\ recognized\end{tabular}}\\ \hline
\cite{Dermy2019} &
  \begin{tabular}[c]{@{}l@{}}Collaborative \\ pick and place\end{tabular} &
  \begin{tabular}[c]{@{}l@{}}Gaze vector from\\ head orientation\end{tabular} &
  \begin{tabular}[c]{@{}l@{}}Max.\\ Likelihood,\\ RBF\end{tabular} & 
  \begin{tabular}[c]{@{}l@{}}Action primitive\\(trajectory)\end{tabular}\\ \hline
\cite{Fathaliyan2018} &
  \begin{tabular}[c]{@{}l@{}}Powdered drink\\ preparation\end{tabular} &
  \begin{tabular}[c]{@{}l@{}}Gaze object sequence\\ in sliding window\end{tabular} &
  \begin{tabular}[c]{@{}l@{}}DTW \\ Euclidean\\ Distance\end{tabular} & Subtask \\ \hline
\cite{Huang2016} &
  \begin{tabular}[c]{@{}l@{}}Smoothie \\ preparation\end{tabular} &
  \begin{tabular}[c]{@{}l@{}}Nr. of glances,\\ duration of 1st fixation,\\ total dwell time, \\ if last glanced item\end{tabular} &
  SVM & Intended ingredient \\ \hline
\cite{Ravichandar2018} &
  Reaching &
  \begin{tabular}[c]{@{}l@{}}Gaze map \\ from head orientation\end{tabular} &
  NN & \begin{tabular}[c]{@{}l@{}}Goal location \\of trajectory\end{tabular} \\ \hline
\cite{Sakita2004} &
  LEGO assembly &
  \begin{tabular}[c]{@{}l@{}}Fixation time, \\ fixations on functional\\ points\end{tabular} &
  \begin{tabular}[c]{@{}l@{}}rule-based,\\ voting scheme\end{tabular} & Type of assembly\\ \hline
\cite{Schydlo2018} &
  \begin{tabular}[c]{@{}l@{}}Placing, giving, \\ eating, pouring\end{tabular} &
  Gaze location &
  LSTM & Action and location\\ \hline
\end{tabular}}
\caption{Approaches to task and intention recognition in human-robot cooperation}
\label{tab:HRI}
\end{table}

Some approaches have used both gaze features and body motion to infer intentions.
\citeN{Dermy2019} used the head/gaze orientation (as estimated from the robot's cameras) and/or kinesthetic help to infer which motion primitive the human is expecting the robot to execute. 
Similarly to \citeN{Singh2020} (see previous section), \citeN{Ravichandar2018} use gaze information as priors to compute the posterior probability of a number of possible trajectories in human reaching as learned via neural networks. In this work too, yet, the gaze is estimated from an external camera mostly relying on head orientation.

Considering action sequences of different lengths, \citeN{Schydlo2018} proposed an architecture relying on gaze and body pose features fed to an encoder-decoder LSTM network, with the decoder layer expanding the K most probable action sequences. Training and testing were done on available datasets containing either placing and handing actions \citep{Duarte2018} or eating and pouring action sequences.  Gaze features improved the results obtained with only body pose, both in terms of accuracy and earliness of prediction.



The approaches reviewed for Human-robot cooperation are summed up in Table \ref{tab:HRI}.

\subsubsection{Shared autonomy grasping systems: assistive technologies and smart teleoperation}
\label{subsec:SA}

Robotic teleoperation in manipulation tasks aims at separating perception and action, by displacing the effects of motor action in a remote position or environment, still exploiting the precise eye-hand coordination of the operator.
In recent years a number of intention estimation approaches have been developed for shared autonomy systems, specifically targeting assistive and teleoperation scenarios.
Shared autonomy combines robotic and human control in the execution of remote tasks. Each partner is then tasked with the aspects for which they are better skilled: the lower kinematic aspects of action execution are usually left to the robot while higher-level cognitive skills, like task planning and handling unexpected events, are typically concurrently exercised by the human. This can happen with different degrees of autonomy for the robotic partner \citep{Goodrich2013,Beer2014,Schilling2016}. Often there is a large asymmetry in terms of degrees of freedom or kinematic capabilities between the user input controller (such as a joystick) and the robotic effector, thus, shared autonomy can facilitate control, in so easing the operator's cognitive load and speeding up execution. 
The user is setting the goals and the ways to achieve them, hence this cooperative interaction cannot be brought about without the robot first inferring the current human intention (\textit{intent prediction}) and only then deciding how much assisting with execution (\textit{arbitration}).\\
\citeN{Losey2018} review a large literature on shared control for physical HRI, specifically in the context of rehabilitation. The unifying view they propose in order to look at any system sharing task execution with a human counterpart states that the design of any system physically assisting a human should take care of three aspects:

\begin{itemize}
\item\textbf{Intention Recognition.} Intention recognition should happen as early and as naturally as possible in order for the user to be relieved of explicitly directing the robot and for the robot to timely initiate the assisting action. A number of approaches have been proposed that rely on \textit{predict-then-act} policies \citep[e.g.,][]{Yu2005,Dragan2013}, predicting one goal and executing a corresponding plan, but a more flexible solution would compute a distribution over all possible goals and rely on continuous reassessment of the predicted intention as the action unfolds, in so providing on line correction and continuous adjustment \citep{Hauser2013,Javdani2015}. 

\item\textbf{Arbitration.} Following intent detection, a policy must be designed to arbitrate {to which degree the user input  and the robot assistance are merged} and executed. This can rely on a predict-then-blend policy,
namely $ Arbitration = (1-\alpha)\cdot UserInput - \alpha\cdot RobotPrediction, \ \ \alpha \in [0,1]$ 
 \cite[e.g., see ][]{Dragan2013}, or a smoother assistance policy computing a probabilistic distribution over all possible user's goals and  minimizing the cost of achieving the user's goal 
   \cite{Javdani2015,Javdani2018}.
\item\textbf{Communication/feedback.} In most scenarios, a robotic system is substituting the motor functions of a human arm and in some cases, its sensors are also used to communicate back to the user properties of the environment.
\end{itemize}
\begin{table}[]
{
\begin{tabular}{|p{2.5cm}|l|l|l|l|}
\hline
\textbf{Study} &
  \textbf{Task} &
  \textbf{Features} &
  \textbf{\begin{tabular}[c]{@{}l@{}}Recognition \\ method\end{tabular}}
  & \textbf{What is recognized}\\ \hline
\cite{Admoni2016} &
  \begin{tabular}[c]{@{}l@{}}Pouring\end{tabular} &
  \begin{tabular}[c]{@{}l@{}}Gaze distribution\\ on AOIs\end{tabular} &
  \begin{tabular}[c]{@{}l@{}}POMDP\end{tabular} & Grasp target\\ \hline
\cite{Razin2017} &
  \begin{tabular}[c]{@{}l@{}}Pick and\\ place\end{tabular} &
  \begin{tabular}[c]{@{}l@{}}Gaze location\\ and velocity\end{tabular} &
  \begin{tabular}[c]{@{}l@{}}LDA, \\ decision trees,\\ k-NN, HMM\end{tabular} &
  \begin{tabular}[c]{@{}l@{}}Action ('move', \\'ready', 'hold')\end{tabular} \\ \hline
\cite{Webb2016} &
  \begin{tabular}[c]{@{}l@{}}Grasping of a\\ tennis ball\end{tabular} &
  \begin{tabular}[c]{@{}l@{}}Gaze location \\ and motion commands\end{tabular} &
  Potential fields & Target location \\ \hline
\cite{Li2017} &
  Grasping &
  \begin{tabular}[c]{@{}l@{}}Gaze 3D \\ location\end{tabular} &
  GMM &  \begin{tabular}[c]{@{}l@{}}Grasp target and \\related contact point\end{tabular}\\ \hline
\cite{Li2020} &
\begin{tabular}[c]{@{}l@{}}
  Grasping for\\ manipulation\end{tabular} &
  \begin{tabular}[c]{@{}l@{}}Gaze and \\ motion\end{tabular} &
  \begin{tabular}[c]{@{}l@{}}BN, SVM,\\NN \end{tabular} &
  \begin{tabular}[c]{@{}l@{}}Grasp target and\\ next manipulation\\ (transfer, usage,\\ handover) \end{tabular}\\ \hline
\cite{Shafti2019} &
  \begin{tabular}[c]{@{}l@{}}Pick and place, \\ pouring\end{tabular} &
  Gaze dwell time on object &
  FSM & \begin{tabular}[c]{@{}l@{}}Primitive action \\ (reach, grasp,\\ pour) + object\end{tabular}  \\ \hline
  \cite{Jain2018,Jain2019} &
  \begin{tabular}[c]{@{}l@{}}Pick and place \end{tabular} &
  / &
  Bayesian filtering & Target object \\ \hline
  \cite{Wang2020} &
  \begin{tabular}[c]{@{}l@{}}Make instant coffee, \\ make powdered\\ drink, prepare cleaning\\ sponge\end{tabular} &
  \begin{tabular}[c]{@{}l@{}}Current AOI,\\ AOI sequence,\\gaze-object angle/speed \end{tabular} &
  RNN & \begin{tabular}[c]{@{}l@{}}Action primitive,\\  target object \end{tabular}\\ \hline
  \cite{Fuchs2021} &
  \begin{tabular}[c]{@{}l@{}}Pick and place\end{tabular} &
  \begin{tabular}[c]{@{}l@{}}Time series of \\ gaze distribution on AOIs \end{tabular} &
  GHMMs &  \begin{tabular}[c]{@{}l@{}}Action and\\  target object \end{tabular}\\ \hline
\end{tabular}}
\caption{Approaches to intention recognition in shared control systems}
\label{tab:HRI2}
\end{table}

In the light of these premises, the possibility to infer manipulation intentions via gaze tracking has become an utmost appealing approach especially in the context of assistive technologies. Here, the intended users are people with physical impairments or elderly for which full teleoperation with a joystick controller is a further challenge. Similar issues to some degree exist also in complex teleoperation scenarios, where the user needs to exert dexterous control in cluttered and possibly dangerous environments (e.g., handling nuclear waste or bomb defusing). Indeed, even in cases where the own hand is used to control a similarly anthropomorphic manipulator, there are low-level kinematic aspects that the robotic part can take care of, in stead of the user, if the right intention can be timely predicted. For example, the best grasp and grasping location can be automatically selected by the robot depending on the current hardware and the target object.


{
Some studies first looked at the theoretical formulation of gaze-based intention estimation.}
As a first attempt at integrating gaze input from the user, \citeN{Admoni2016} presented a proposal relying on Javdani's framework \citep{Javdani2015} where the probability distribution over the goals (hidden states) is updated  by considering both user's eye movements and joystick commands as observations in a POMDP, using hindsight optimization to solve it in real time. 
Assuming the position of the manipulated objects to be known (by using fiducial markers), the probability of a certain goal given the current image is considered proportional to the distance between the current gaze position and the target object. {Still, in this study no experimental data or a practical implementation was presented.}

In a more typical shared autonomy context,
\citeN{Jain2018} put forward a mathematical formulation of intent recognition in a recursive Bayesian filtering framework. Assuming a finite number of available goals, the probability distribution over these is computed as the posterior conditioned on a number of observations (gaze, motion input commands,...) and depending on the previous goal estimation in a Markov model. The MAP is then considered as the current goal, with a confidence dependent on the difference between the first and the second most likely goal. 
The amount of assistance provided by the system is regulated by the confidence in goal prediction. 
Overall, this study shows also the importance of identifying the right trade-off between early assistance and high confidence in the predicted user's goal.

In a further work \citeN{Jain2019}, the same mathematical formulation is extended with a user-customized optimization of the rationality model, in so accounting for human action suboptimality. 
Still, in both works gaze is mentioned as a possible input to the system but not practically integrated and tested.\\

Most shared autonomy approaches relying on eye tracking for intention recognition thus far are rather recent and often do not entail all three components presented above. {Such systems usually} focus on the first component or subcomponents of it, like gaze 3D localization or investigation  of gaze patterns in shared autonomy contexts.
{The following studies considered the characterization of gaze behavior during teleoperation  and how it could be learned for intention estimation.}

For an assistive robot arm spearing food bits from a plate to feed an impaired user, \citeN{Aronson2018a} presented a preliminary eye-tracking experiment comparing user behavior within-subjects in different operation modalities: full teleoperation, complete autonomy, shared autonomy according to \citeN{Javdani2015}, blend-policy autonomy as in \citeN{Dragan2013}. Two patterns of fixations were observed: monitoring glances, meant to check the translational behavior of the arm approaching the intended food morsel, and planning glances, which select the target morsel before starting the arm actuation, as in natural eye-hand coordination \citep{Johansson2001,Hayhoe2003}. 
Yet, an effect of operation condition on the frequency of this kind of fixations was not found 
{but some} patterns of repeated monitoring glances were observed, as the gaze was switching between the effector and the target morsel. 
Still, eye data were further not really localized with respect to the scene content, and analyses were focused just on gaze dynamics, hence limiting the conclusiveness of some results. The lack of localization of the gaze on the object of interest and the possible drop in data quality was further addressed in a subsequent study \citep{Aronson2019}, with semantic labeling, 
while the same group also proposed to use gaze tracking for error detection during manipulation \citep{Aronson2018b}. 
3D gaze localization and its use in intention prediction was investigated by \citeN{Li2017}. In this case a visuomotor grasping model also predicts how a user intends to grasp the chosen object. This latter feature still is formalized for just one cuboid object, resulting in the inference of one contact point and direction of approach of the hand, depending on where the cuboid was fixated, as learnt from a subject physically grasping the object in a training phase.
Recently, relying on the approach first put forward by \citeN{Fathaliyan2018}, \citeN{Wang2020} proposed a framework for action primitive recognition by means of 4 features: the currently gazed object, a gaze object sequence complemented with the gaze object angle (angular distance between the gaze vector and the vector to each object) and the gaze object angular speed. Four action primitives (reach, move, set down, manipulate) were envisioned in the context of 3 activities (make instant coffee, prepare a powdered drink, prepare a sponge for cleaning). The proposed system based on a recurrent neural network achieves good recognition accuracy and observational latency, that is a correct prediction was available on average after 10\% of the action was observed. Although aimed at shared autonomy in assistive contexts, the system  was trained on natural eye-hand coordination, which in the actual scenario might be strongly mediated and influenced by the visual input and by the robotic actuator.
\citeN{Fuchs2021} partially addressed this issue by presenting the scene in a VR headset and letting the user teleoperate a rendered robotic hand by means of natural arm movements (tracked via Vive controllers). The sequences of gazed AOIs during a pick-and-place task was used to train two action-specific HMMs able to generalize their accuracy to new users, number of objects and object configurations. Still the system was not yet tested in a shared control scenario.
\\

 {As in the previous section, a number of studies combined gaze and motion features, in the effort to improve prediction and increase robustness.}
 \citeN{Razin2017} tested a number of modalities, features, and classification techniques to predict intent during robotic eye-hand coordination. Hand motion, eye, and gaze features are used to predict 3 tasks ('move','ready to move', and 'hold') in a scenario where a robotic claw is used for pick and place tasks.  In such conditions, not surprisingly, the authors find that motion features suffice to predict the three tasks, with eye and gaze features achieving good levels of prediction but yet not improving performance when considered together with motion. Decision trees, k-Nearest Neighbor and Hidden Markov Models achieved the best performance in terms of average accuracy and Area-Under-the-Curve (AUC). {This study yet aimed at distinguishing just the motor action, without the object of interest. Still, when a cluttered scene is presented to the user, the gaze provides a more precise and earlier information with respect to motion trajectory features, which in the beginning might be compatible with multiple targets, as shown by \citeN{Aronson2021} for motor control with a joystick and  by \citeN{Belardinelli2022} for natural movement control in VR.}\\

{In the works reviewed thus fare, only the immediate action and object of interest are predicted.}

In a more recent study \citeN{Li2020} {considered} whether besides the target object also the  manipulation intent can be recognized before the reaching motion is completed, in order for the robot to select the best grasp for the following task. In this study, hand tracking is used to control a robotic arm with a three-finger gripper. Since the same grasp could lead to different tasks (and sometimes the same task is still executed with different grasps) the whole problem is treated in a multi-label classification framework. The intent inference is modelled via Bayesian Network (and for comparison SVM and NN) computing the MAP likelihood of an intent given an object, on which the gaze is lingering, and a grasping configuration. Experiments were conducted with 4 objects and 3 tasks, with the training set collected in a human grasping experiment and testing conducted on both real grasping sequences and teleoperated sequences. 
{Again,} unlike in \citeN{Razin2017}, gaze information seems to increase performance in the teloperated case, less so in the real grasping case. It must be noted that a shared control was not yet implemented in this framework.\\

Other systems present some form of intent detection already integrated in a shared autonomy framework.
\citeN{Webb2016} consider a remote teleoperation scenario and a hybrid joystick-gaze control strategy to overcome the hampered depth perception when manipulating objects via screen feedback, by restoring a kind of more natural eye-hand coordination. 'Look-ahead' fixations are used to predict the target of the movement and a Gaussian potential field is used to combine this input with the direct joystick commands, in a predict-then-blend framework. This strategy resulted in fewer grasp attempts and shorter execution times and was more liked by the participants compared to a joystick-only control.
Yet, in this system, the gaze was actively used to indicate the target. To discriminate between voluntary and involuntary gaze shifts a button press on the joystick was required when looking at the target object, while the grasping was also triggered via button press, reducing the claim of natural eye-hand coordination. 
One of the few  examples where gaze-based intention recognition is integrated in a more elaborate assistive framework for rehabilitation is presented in \citeN{Shafti2019}. The system here relies on 3D gaze localization, convolutional neural networks for object recognition, head tracking, and a finite state machine (FSM) implementing a basic action grammar according to the predefined 'graspability' and 'pourability' of the objects in the scene. Indeed, besides pick and place on a table, also pick and place in a container and pour into a container were chosen as tasks. Here too, the user is required to actively fixate the object on the right side to signal an interaction intention, while the robotic arm attached to the user's arm carries out the action. 

{The studies presented here and summarised in Table \ref{tab:HRI2} show the fast increasing interest in using gaze in assistive HRI and teleoperation, but, more so than in other fields, also show a number of critical limitations and challenges that will be further addressed in Section \ref{sec:issues}.}

\subsection{Inferring driver's intentions in ADAS}
\label{subsec:ADAS}
Advanced Driving Assistance Systems (ADAS) have been developed in the last decades in order to improve driving safety, not just by protecting passengers in case of an accident, like passive systems (seat belts, airbags, ...) do, but by preemptively decreasing accident risks \citep{Koesdwiady2017}. {Until autonomous driving becomes commonplace in multiple contexts}, it is important to also invest in systems that support the driver at any automation level less than level 5. As we have seen, driving is a rather long and complex behavior that assumes a number of high-level cognitive capabilities (sustained attention, multi-tasking, situation awareness, decision making, eye-hand-foot coordination,...) to constantly work at full capacity. Even if part of this behavior is somewhat automatized, the cognitive load in some situations can be very high and the smallest amount of fatigue or inattention can have fatal consequences. In this context, a number of solutions have been explored: on the one hand, many research efforts consider gaze tracking to assess the driver's cognitive state and possibly warn him if they get distracted \citep[e.g.,][]{Ahlstrom2013}; on the other hand, especially in the context of semi-autonomous driving (up to level 3), efforts are being directed at assisting in the execution of difficult maneuvers. In this latter case, of course, the earliness of the intention estimation output could be critical, since the time range for the ADAS to act on this estimation is even shorter than in teleoperation, especially considering vehicle speed compared to human motion speed. This also depends on what the ADAS needs to do with information about the driver intention: it can just warn (by accounting for human reaction time), assist (by planning and executing a supporting plan) or simply avoid taking decisions conflicting with the driver's intent, like triggering a lane departure warning when a lane change is indeed intended. \citep{Xing2019}.\\


Although the idea of using gaze to infer driving intentions dates back to a couple of decades ago \citep[e.g, in ][where up to 7 maneuvers were recognized via HMMs]{Oliver2000}, in the following years mostly other behavioral cues have been used for intention estimation in this context, due to the technical difficulties of performing eye tracking in a car, with changing light conditions and a moving scene.

In particular, Trivedi and colleagues have long worked on driver's intent estimation, yet they mostly used head dynamics data, showing that in lane changing this provides a more significantly discriminative cue than gaze data \cite{McCall2007,Doshi2009}.

\citeN{Liebner2013} used the head heading angles coupled with vehicle velocity and indicator activation to infer sequences of intentions like lane changing and turning. 

{Focusing on gaze,}  \citeN{Lethaus2013} compared a neural network, a Bayesian network, and a Naive Bayes classifier in discriminating a lane change versus lane keeping (or lane change right versus left). By collecting gaze data in a driving simulator, the authors looked both at the amount of data necessary for the model to make a prediction and how soon a prediction could be made before the maneuver started. Gaze time on 5 zones (rear mirror, left/right mirrors, windscreen, and speedometer), in time windows of 10 and 5 s before the lane change, was considered as input for the models. As to be expected, the closer the prediction to the maneuver start, the more accurate the estimate for all models. A 5s window turned out to produce better predictive models since less noise from other activities was included, with neural networks delivering slightly better results than Bayesian Networks.
\citeN{Peng2015} found a similar window of 5 s second necessary when using head rotation standard deviation along with other features concerning the vehicle motion, the state of the indicator, and lane changing conditions.

\cite{Wu2019} considered intention estimation within a semi-autonomous vehicle. In this context, if the system is uncertain about the best maneuver when following a lead vehicle that starts decelerating, it should alert the driver to take control. To estimate the driver's intention to brake or change lane just by means of gaze/fixation data, the authors designed an HMM coupled with probabilistic Dynamic Time Warping to model the probability of each new observation given last observations and the last estimated intention in a recursive Bayesian estimation scheme. An accuracy of over 90\% for the 3 intentions could be demonstrated with an anticipation time over 3s before the maneuver actually started.

While most intention estimation systems in ADAS deal with the prediction of lane change and braking, \cite{Jiang2018} consider a different application: here the user is supposed to be a passenger in an autonomous vehicle and the system is trying to retrieve their point of interest, such as shop signs or a location the user intends to navigate to, as the vehicle moves, to possibly offer related information. 
By using Markov Random Fields, the proposed solution estimates dynamic interest points and shows robustness to noise sources due to eye tracking inaccuracy as well as to the moving scene and the vehicle bumpy motion. Still, in this case, as in \citep{Huang2016}, not a precise motor intention is retrieved, rather an object of interest.

{As the studies presented in this section show,  although intention estimation is critical for ADAS,} gaze-based intent prediction is still in a rather early phase, indeed in such scenarios  where prediction errors might have fatal consequences the use, even for warning systems, is still very moderate.
{
\section{Current limitations, issues and open challenges}}
\label{sec:issues}

{The literature surveyed in the previous sections shows that scientific research and technical progress on gaze-based intention estimation have reached a consistent level of maturity to let foresee forthcoming application in technological products, especially in cognitive robots. Still, even though eyetracking capabilities are increasingly embedded in multiple interfaces, the same speed has not been witnessed for robust algorithms leveraging eye movements. As seen, applications and solutions have usually a limited scope and rely on carefully designed supervised training. While we hope that a better understanding of the cognitive underpinnings might help with that, it is also the case that certain limitations derive from open issues and challenges both in the current understanding of how the visuomotor system serves the schema control and how technical systems harness eye movement data.
Further, although perhaps not central to the mere functioning} of an intention prediction framework, there are still other issues and themes that can be critical for the design of a user-friendly and effective system, one which the user is also able to trust and hence willing to interact often with. These issues can indeed also influence the user's gaze behavior and should be taken into account. We present here some of the most relevant ones, along with positions and approaches that have been put forward in the literature.\\

{
 \subsection{Multitasking nature of gaze, gaze as a proxy of attention, and bimanual manipulations}
As already mentioned, at any time multiple cognitive processes are ongoing in our brains. This is especially true when considering more complex activities than grasping or selecting items, like during driving or carrying out a manipulation task made of multiple steps in a loosely constrained sequence. Already \citeN{Land1999} recognized how fixations on objects could be  categorized according to four different functions (locating, guiding, directing, and checking), yet these could also be  interleaved with fixations from different subtasks: so even though the eyes are monitoring the current execution some fixations could proactively target already the next object to check its location  and then go back to the current one. This kind of scheduling is rather inscrutable currently and has high variability across contexts, tasks, and people. \\
On the one hand, if eye fixations can reveal a lot about the information accrual process and are usually considered as a behavioural proxy for the attentional focus, it is at times an unwarranted assumption that everything in a fixated region is consciously detected. It is, for example, easy to experience time gaps when driving, when our internal focus is directed to unrelated thoughts and still the driving and visual scanning behaviour can proceed normally, according to automatized patterns.
On the other hand, especially when controlling multiple effectors, some monitoring could be done covertly by peripheral vision or by strategically fixating some middle point, hence not hitting relevant AOIs. In natural bimanual eye-hand coordination, we usually visually monitor more closely the dominant effector, executing the most difficult action and rely on covert attention, proprioception and haptic feedback to control the other limb \cite{Srinivasan2010}. Alternatively, we shift our eyes between the two targets in an interleaved fashion \cite{Riek2003}. However also this kind of scheduling is not well researched, especially across people and handedness. \\
Most current gaze-based systems in robotics, to our knowledge, deal with the control of just one arm, while recent bimanual
frameworks in shared control rely on learning directly both
hand motion patterns or the second robotic arm is
autonomously coordinated with the motion of the operator’s
right arm \cite{Rakita2019,Laghi2018}. That is, gaze-based intention estimation for multi-tasking and multiple effectors remains still an open problem, especially when it is difficult to discern what has been covertly perceived or altogether overlooked.}

\subsection{Bidirectional effects on users' gaze behaviour}

In Human-Computer Interaction, gaze-based control of an input device has been limitedly used and mostly in assistive contexts \citep[e.g.][]{Zeng2020}, that is, in the case of patients with impaired limb control. A gaze-based controller indeed can be at times problematic because of the "Midas touch" effect \citep{Jacob1991}. We are in fact not used to control our eyes as effectors, but to use them for perception in a mostly automatic way, without producing intentional effects in the physical world. {Thus,} gaze-based selection in such interfaces can lead to a number of false selections. Indeed, distinguishing glances to perceive and glances to act becomes difficult for the system, while it can lead the user to very controlled, unnatural eye movements, in turn increasing the cognitive load. This might be less the case in a gaze-based intention estimation system, still: if the user becomes aware that the system is using their gaze to produce an assisting effect, they might modify their natural scanpaths to actively obtain the intended effect.  As a result, scan patterns would diverge from the behavior seen during training without assistance, consequently throwing off the system. The general assumption, as in \citeN{Javdani2015},  is in fact that the user is not minding the system while pursuing their own plan. If instead, the user learns that their gaze has an effect on the robot motion, they might actively use it as a further control input to guide the robot: this might be on purpose so or just a side effect that needs to be considered. For example, in a driving context, if users learned that failing to look at the mirror when changing lanes would trigger a warning, they might actively plan a gaze check to the mirror to avoid the nuisance, triggering instead the lane change assistance.\\

In teleoperated manipulation settings, the visuomotor task needs first to be learned, by acquiring the novel mapping between own movements and the distant visual sensory consequences. This implies that until a ceiling level of training is reached, gaze patterns change along with training. \citeN{Sailer2005} investigated how the gaze behavior evolved when learning to control a cursor on a screen for an intercept task by means of an unfamiliar device applying isometric forces and torques. They observed a first exploratory stage where the gaze was mostly pursuing the controlled effector (the cursor) and the hit rate was rather low. In the second stage, the eyes became more predictive of the striven cursor position, while the hit rate improved. Finally, once the visuomotor task was mastered, the eyes would jump directly to the target, without the need to monitor the cursor movement.  While this was a rather unfamiliar task, it could be possibly comparable to controlling a robotic limb by means of some input device utterly different than the own hand motion. 

Effects on scanpaths while controlling a robotic arm have recently been discussed by \citeN{Aronson2018a, Fuchs2021,Wang2020},  but further investigation will be needed to shed more light on the way robot-mediated eye-hand coordination might influence how healthy and disabled users manage to orchestrate perception and action processes. 
Such an issue is in general also related to user adaptation and the the degree of trust in the system, as discussed next.
{\subsection{Adaptation, co-adaptation and trusting the assistance system}}
{
Considering predict-then-blend approaches in shared autonomy, there is usually an arbitration parameter $\alpha$  determining how strongly the robot assistance will be implemented with respect to the user's input. Small values can determine too timid assistance, hence not really supporting, while too large values can lead to aggressive assistance, which can be annoying \citep{Dragan2013}. The amount of assistance, of course, should depend also on the confidence in the inferred intention, to avoid aggressively assisting with the wrong goal.
To adapt arbitration to the user and hence find the optimal amount of assistance, \citeN{Oh2019} propose to learn the best arbitration parameter online from the user interaction with the system.
Considering in general human-machine interfaces, \citeN{Gallina2015} proposed an approach in between goal-oriented design (which maximizes performance) and human-centered design (which maximizes usability): \textit{progressive co-adaptation}. The idea here is that as the user through interaction with the system adapts and improves their skills , concurrently the system should monitor these improvements and adjust its parameters accordingly. As shown by \citeN{Sailer2005}, this learning has also an effect on gaze, which should be taken into account. Yet, in most HRI approaches a model is learned by collecting data during the targeted task and usually not retrained in light of long-term use and adaptation by the user.   Again this is an open problem and a possible solution might entail online learning to adapt the parameters of the time series prediction model as the data distribution changes \cite[e.g., ][]{Guo2016}. }



While \citeN{Javdani2015} make the assumption that the user is agnostic as to the robot assistance and does not consider it in its control strategy,
recently \citeN{Amirshirzad2019} investigated more systematically instead exactly how humans adapt in teleoperation versus shared control. Performance sharply improves in the latter condition compared to the former, after an initial exploration phase, where users implicitly learn the robot behavior and adapt to it. 

Finally, a further aspect to consider is that,
as suggested by \citeN{Fathaliyan2018} and  \citeN{Aronson2018b}, by means of gaze also the degree of trust in the system, anomalies due to malfunctioning, or other issues can be assessed. Indeed, recalling the results of \citeN{Sailer2005}, in teleoperation scenarios, for example, once training is completed and the user is familiarized with the system, less fixations will be placed on the robot and the gaze will become more predictive of the targeted objects/locations. Indeed, brief visual assessment of the success of the current subtask will be followed by a gaze shift to acquire information for the following subtask \citep[see e.g., ][]{Fuchs2021}. This will probably not be the case if the user is constantly monitoring the robot to make sure that it does its part. On the other hand, if after some time of trusted use of the system the gaze deviates from the usual predictability, it can be a reliable sign that something either in the system or in the mental state of the user has changed. Such hypotheses of course will need suitably designed user studies and the collection of both objective and subjective measures to test performance, user satisfaction, and experience.\\

\subsection{Trade-off sense of agency/speed of execution: the case for personalization}

Letting the user retain their sense of agency is a crucial aspect for most human-machine interaction systems, even when control is shared transparently and the machine can correctly read the user's intention. Also in semi-autonomous vehicles, indeed, if the human driver takes a decision and executes a maneuver, an override from the system might substantially damage the trust in the system.

In \cite{Pacherie2008}, the sense of agency is defined as the sense of intentional causation and it manifests in the subjective perception of a match between the prior intention and the following action \citep{Wegner2005}. This is strongly related to the ability to predict the consequences of our actions. If an artificial system interferes with this prediction and produces an action execution that does not match our expectations, we might feel we are not the agent causing the action. This might be sometimes at odds with the assistance the system is trying to deliver. {Further, it will cause eye movement patterns to change, becoming again less anticipative. Still, behavioral measures for such sense, and in particulat oculomotor measures, have been hard to identify \cite{Schwarz2018}.}

Most works on shared autonomy systems, for example, report user studies with objective and subjective measures.
If task completion times usually benefit from human-robot cooperation, subjective opinions give a less clear-cut picture.
Sometimes for complex tasks more autonomy is preferred, while for other simple tasks more control to the human is preferred.
\citeN{Dragan2013} found that these preferences nonetheless interact with other factors, like the arbitration style (timid vs. aggressive) and the intention prediction accuracy (right vs. wrong), and in general there are also subjective differences in these preferences.

{In this context, contrasting results which found a preference for execution speed \citep{Dragan2013,Hauser2013} suggest} that the different ways participants interact with the system call for some personalization of the cost function modeling the user.

\citeN{Gopinath2016} picked up on this suggestion, noting that in the assistive context, users differ in their motor capabilities (and these can even both deteriorate or improve) and might need different degrees of assistance. {An approach grounded in optimal control theory is proposed, 
with the tuning of the arbitration parameters  formulated as an optimal control problem}, where the user's and robot's control signals are given. The optimization is carried out by the users themselves, increasing or decreasing assistance by verbal commands. 
{While tasking the user with a further parameter to control in another input modality,} the study still paves the way for a more thorough investigation of personalization within shared autonomy systems, with the aim of optimizing the balance between user satisfaction and system effectiveness. 

This is still not just a matter of arbitration as individual differences in scanpath behavior might call for personalized training also in intention estimation \citep{Wiebel2020}. In simple motor tasks like pick and place scanpath patterns might look very similar across users, while in less constrained, more complex activities the oculomotor behavior might be more directed according to the way each person schedules the subtasks \citep{Yi2009}. 
{
\subsection{Extracting more information from 3D gaze data}
As seen, collecting quality gaze data in 3D and localizing them on real-word objects, is still a matter of research (see Section \ref{subsec:HRI}). 
As argued in Section \ref{subsec:eye-hand}, also the type of grasp and further manipulation intent 
in some cases can be inferred from fixations. Still, these aspects have been hardly 
tackled in technical systems, because it is often computationally hard to recognize 3D objects in
any pose and appearance, while also localizing online the gaze with head motion on their functional parts without fiducial markers. 
This would also need the definition of multiple \textit{ad-hoc} AOIs for each object 
and infusing in the system task, grasp, and functional knowledge 
about predefined objects. In this sense,  gaze-based intention estimation can achieve better results  when embedded in more advanced cognitive system, not just relying on precise eye-tracking capabilities but also on a sophisticated scene and world understanding, as currently developed social and assistive robots.}



\section{Conclusions}
As stated by \citep{Vernon2016}, intentions are the cognitive bridge integrating goals and actions in a prospective framework. Inferring human intentions is thus crucial for assistive and cooperative systems.
In this survey, methods and applications of gaze-based intention estimation have been reviewed and summarized, along with the underlying cognitive science underpinnings concerning perception and action. Eye tracking capabilities have made large progress in the last decades, allowing moving from constrained and controlled laboratory settings to real-world activities. Concurrently, machine learning, robotics, and ADAS research endeavors have reached a point where integrated cooperation of humans and machines can finally find application in common tasks such as daily living activities, driving, or human-robot cooperation. This interaction can be made more effective and integrated by inferring the human intention based on gazed objects, endowing the robotic partner with an awareness of what the user wants to achieve and letting it assist in a seamless way. In this respect, surveyed studies and frameworks show promising results and perspectives, at least for relatively constrained tasks and applications. They compellingly show how gaze features are crucial for early and accurate inference of intentions and to assess the cognitive state of the user. {Yet, on the one hand, a more principled consideration of the targeted cognitive processes and subsequent oculomotor manifestations, as reviewed here, could improve the design of such systems, especially selecting the most adequate features and prediction models. On the other hand, some open issues and limitations need still to be investigated and improve for intention estimation to achieve wider spread in cognitive systems. }

Important challenges and opportunities are connected to these technologies, especially regarding physical systems. On the one hand, robots that could be operated remotely in a natural way and support the operator by taking care of low-level motor aspects would find wide applications in crises like the pandemic currently afflicting the world. Similarly, assistive systems able to infer and execute actions in peripersonal space would enable people with certain disabilities to act in their environment and regain a certain amount of independence. On the other hand, gaze-based driving assistance systems would support the driver in a smarter and more reliable way, also possibly providing a basis for learning from their perceptual decision making and paving the way for a more human-like full autonomy.

\bibliographystyle{ACM-Reference-Format}
\bibliography{gaze_pred.bib}

\end{document}